\documentclass[10pt,twocolumn,letterpaper]{article}

\usepackage{cvpr}
\usepackage{times}
\usepackage{epsfig}
\usepackage{graphicx}
\usepackage{amsmath}
\usepackage{amssymb}
\usepackage{url}

\cvprfinalcopy 

\def\cvprPaperID{****} 

\begin{document}

\title{A Fast HOG Descriptor Using Lookup Table and Integral Image}

\author{Chunde Huang\\
Xiamen University\\
422 Siming S Rd, Siming Qu, Xiamen, China, 361005\\
{\tt\small xmuhcd@msn.com}
\and
Jiaxiang Huang\\
Xiamen University\\
422 Siming S Rd, Siming Qu, Xiamen, China, 361005\\
}

\maketitle

\begin{abstract}
   The histogram of oriented gradients (HOG)\cite{dalal2005histograms} is a widely used feature descriptor in computer vision for the purpose of object detection. In the paper, a modified HOG descriptor is described, it uses a lookup table and the method of integral image\cite{crow1984summed}\cite{huang2009fast} to speed up the detection performance by a factor of 5$\sim$10. By exploiting the special hardware features of a given platform(e.g. a digital signal processor), further improvement can be made to the HOG descriptor in order to have real-time object detection and tracking.
\end{abstract}

\section{Introduction}
The process of using a HOG descriptor to detect object includes feature generation and region scanning. A widely used implementation is given by the OpenCV. It's implemented in a way for general application purpose based on the OpenCV's large programming functions libraries. For a image of size 800$\times$600, it takes 100s ms to process in a common PC (Intel i3/i5/i7 CPU). To speed up the procedure, a lookup table is used to compute the oriented gradients. Afterward the integral image method is used to cache the histogram features, which will be used to predict object bounds and objectness scores at each position. Details of these two modifications will be described in the next two sections.

\section{Lookup Table for Gradients Calculation}
\begin{figure}[htb]
	\begin{center}
		\includegraphics[scale=0.45]{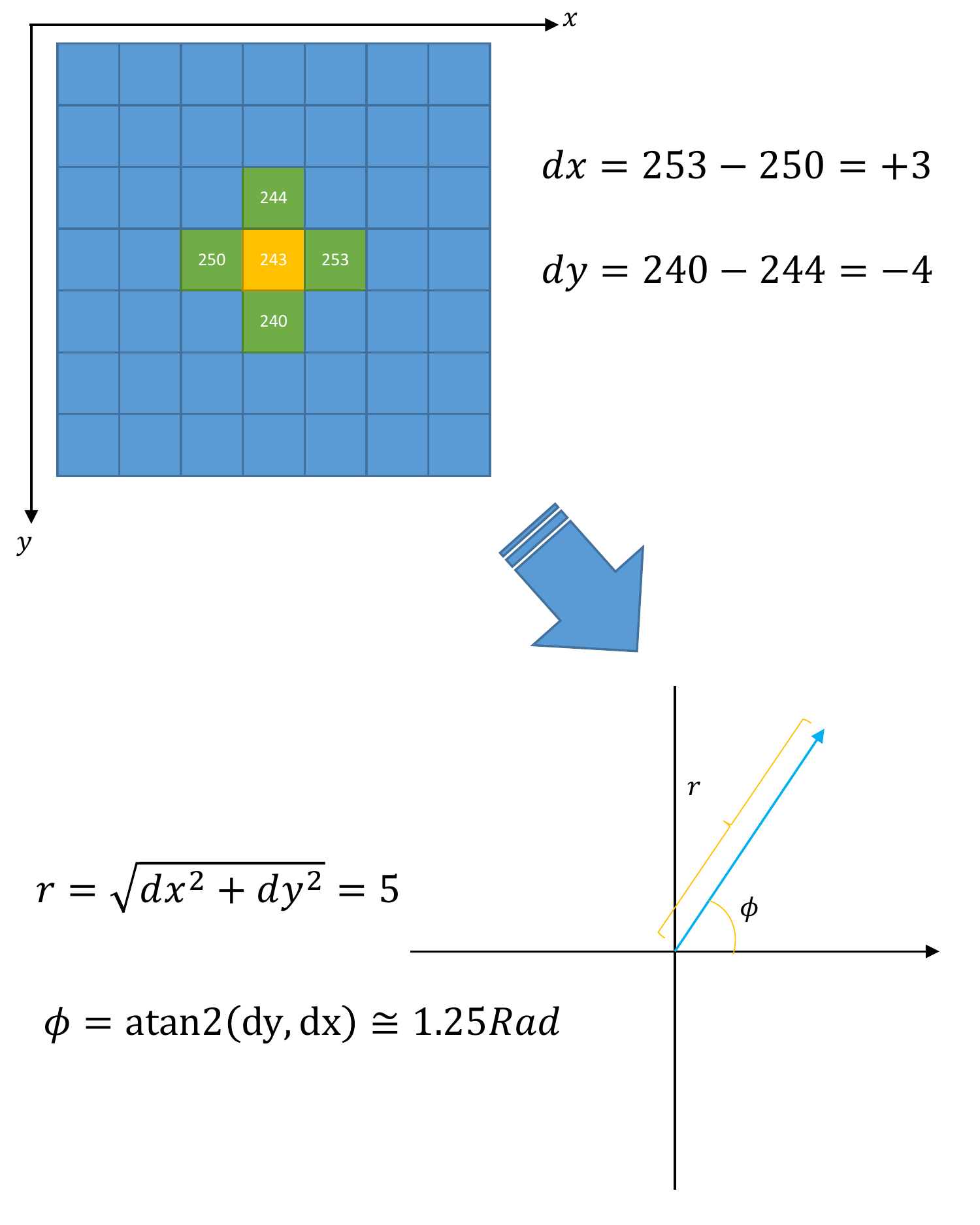}
	\end{center}
	\caption[Calculation of oriented gradients]{The calculation of oriented gradients, it's computationally expensive for a image with over hundred thousands pixels.}
	\label{fig:Cal_HOG}
\end{figure}

The first step to generate HOG feature requires the oriented gradients information at every point for a given image. Let $P_c[x][y]$ be the gray value $(0\sim255)$ for a pixel at position $(x,y)$ of the color channel $c$ $(c=R,G,B)$. To find the oriented gradients of the pixel, the gray value changes in both horizontal(x direction) and vertical (y direction) directions should be computed using the current pixel's previous and next neighborings' gray values. The differences in x and y directions are denoted as:\\
\begin{equation}
\left\{
\begin{array}{ll}
dx=P_c[x+1][y]-P_c[x-1][y]\\
dy=P_c[x][y+1]-P_c[x][y-1]
\end{array}
\right.
\label{eq:ii_HOG}
\end{equation}
\indent Transforming the pair $(dx,dy)$ from Cartesian coordinates to polar coordinates representation yields the magnitude $r$ and orientation angle $\phi$, which are the oriented gradients. Since the $dx$ and the $dy$ are integers ranging from -225 to 255,  a lookup table with $511\times511$ entries for each pair of $(dx,dy)$ can be constructed and fill with precomputed values. Then for every given input image, its oriented gradients map can be built very fast simply by looking up the pre-built table, without doing any computationally expensive coordinates conversion. One example of calculating the oriented gradients is shown in the Fig.\ref{fig:Cal_HOG}. It should be clear that the direct calculations of the magnitude and orientation angle would be much slower than querying a table. Besides, the lookup table needs no more calculation once it's constructed at the very beginning.

\begin{figure}[htb]
	\begin{center}
		\includegraphics[scale=0.35]{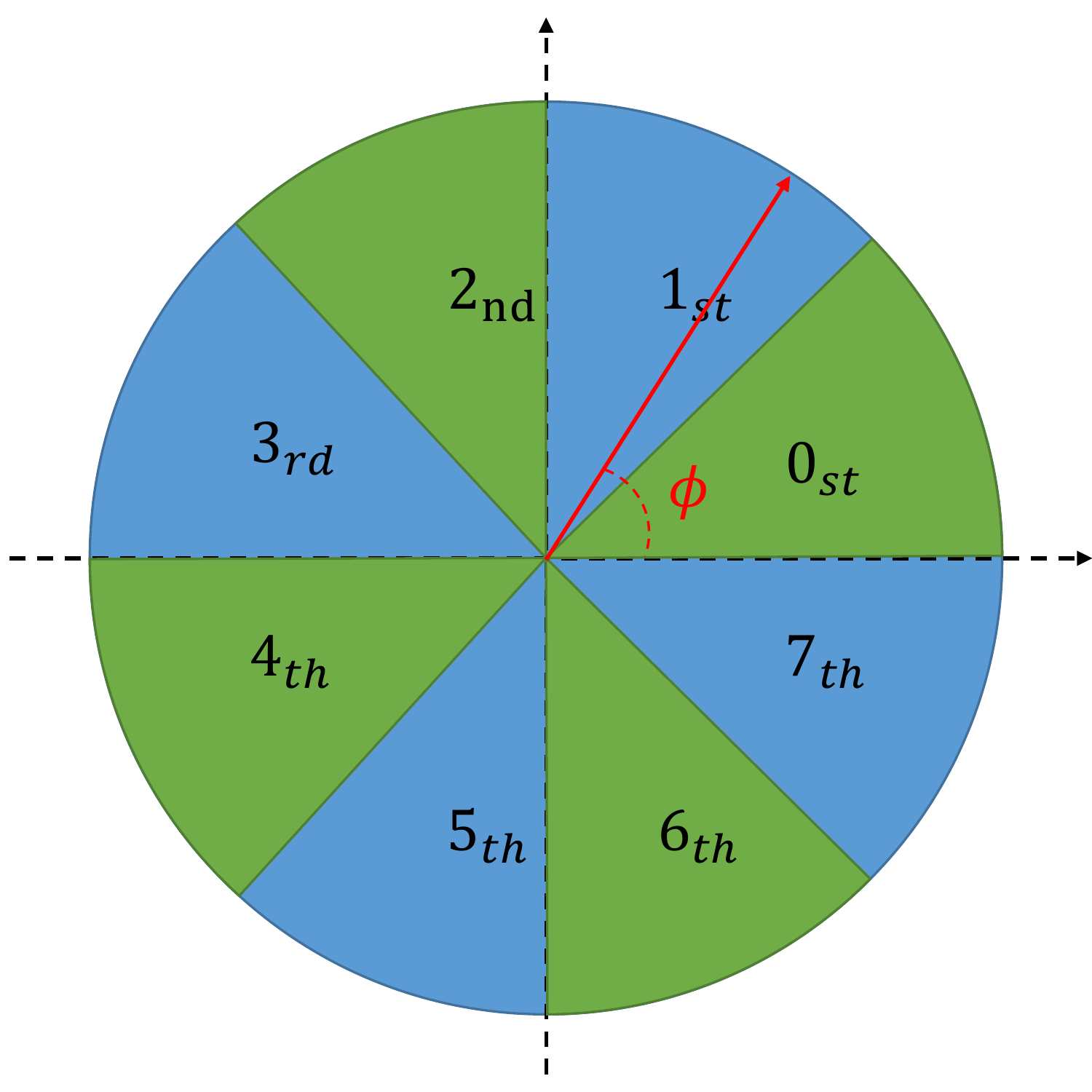}
	\end{center}
	\caption[Discretization of gradient angles]{When compute the HOG feature, the directions of oriented gradients are discretized into bin numbers. Here the $2\pi$ angular space is spitted into 8 areas, each area is called one bin. The oriented gradient $\phi$ falls in the second bin.}
	\label{fig:Bins}
\end{figure}

Additionally, based on our experiments, the magnitudes of gradients are found to be  insignificant for object detections. Thus the magnitude for every point can be simply set to unity when building the oriented gradients map. The angular direction $\phi$ is discretized to an integer bin number, ranging from $0$ to $N_b-1$, where $N_b$ is the number of angular bins, see the Fig.\ref{fig:Bins}. \\
\indent For the simplest method, each pixel for each given image will be associated with a bin number. In the implementation of the OpenCV, each pixel's HOG magnitude part will be distributed to two bins when building histogram features, but this does effect the application of the integral image. In the rest part of this paper, we only focus on the simplest case, i.e, each position of an image only associated with a bin number, and the magnitudes are set unitary.

\section{Application Integral Image}
Integral Image is also known as summed area table, which is commonly used for quickly and efficiently generating the sum of values in a rectangular subset of a grid,it allows for very fast feature evaluation, e.g. an image input for object detection\cite{viola2001rapid}\cite{viola2004robust}\cite{bay2006surf}. In this paper, we only concern 2D grid data since the oriented gradients map is arranged in 2D pattern. 
\subsection{Integral Image Representation}
\begin{figure}[htb]
	\begin{center}
		\includegraphics[scale=0.65]{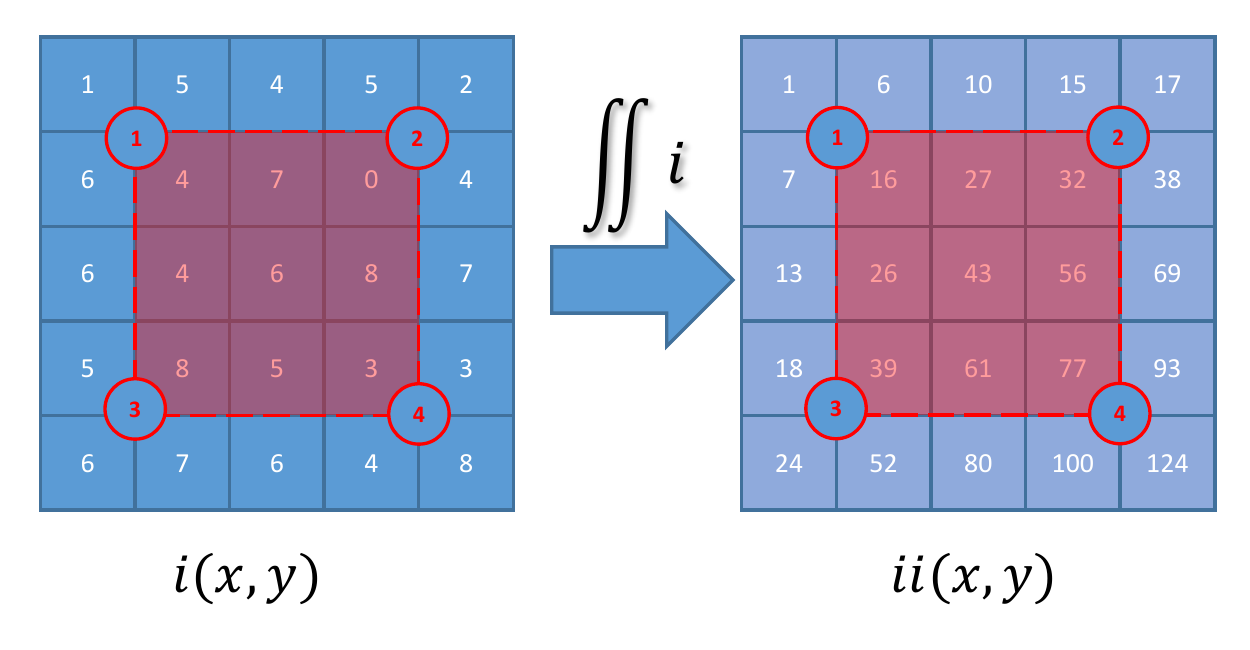}
	\end{center}
	\caption[Discretization of gradient angles]{Calculation of a integral image for a $5\times 5$ grid region. The shield sub region is the area of interested. See the text for more information}
	\label{fig:integralimage}
\end{figure}

Assume the 2D grid data is of size $W\times H$, its value at location $(x,y)$ is $i(x,y)$. Its integral image value at $(x,y)$ is denoted as $ii(x,y)$ , then $ii(x,y)$ can be calculated using the Eq.\ref{eq:ii_cal} An example of doing the integral over a small area is shown in the Fig.\ref{fig:integralimage}\\

\begin{equation}
ii(x,y)=\sum_{x'\leq x,y'\leq y}i(x',y')
\label{eq:ii_cal}
\end{equation}

To calculate the area sum ($V_R$) of an interested region $R$ (e.g, the shaded area in the Fig.\ref{fig:integralimage}),we do not have to add up the value of every single point inside that area. Instead, its integral image representation can be used to compute the sum easily as shown in Eq.\ref{eq:ii_rep}:
\begin{equation}
V_R = ii(4) + ii(1) - ii(2)-ii(3)
\label{eq:ii_rep}
\end{equation}

\begin{figure}[htb]
	\begin{center}
		\includegraphics[scale=0.65]{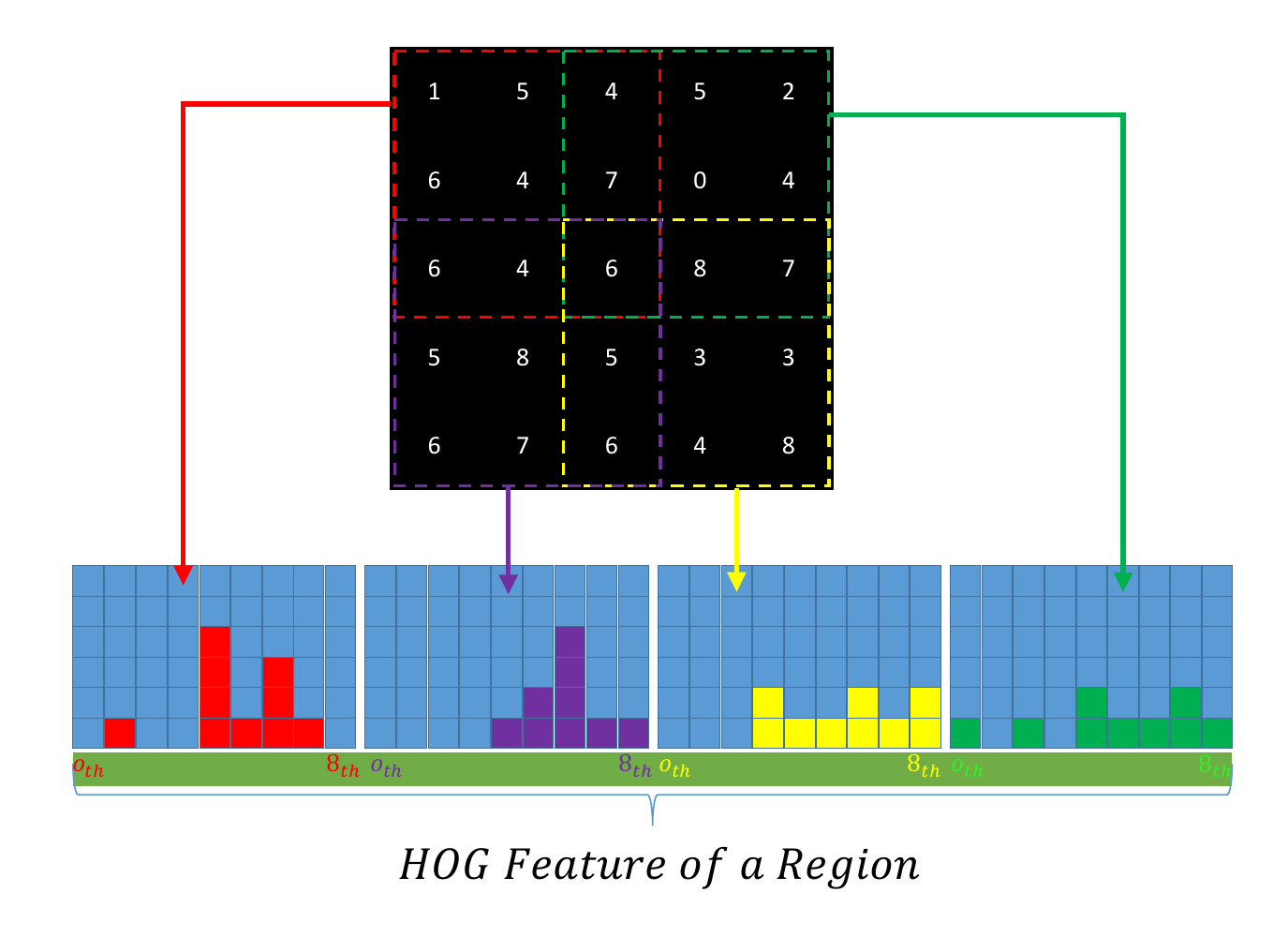}
	\end{center}
	\caption[HOG Calculation]{The HOG features of a interested region is generated by connecting the HOG features of its four sub areas. The four histograms are the features for the four overlapped areas as indicated by dashed rectangles.}
	\label{fig:HOG_Gen}
\end{figure}

An oriented gradients map of an image comprises a 2D grid of bin numbers (for zero-based, these numbers are from $0\sim N_b-1$). This map can be used to build the HOG features for a region inside it. To determine if a region contains object of interested, the HOG features in that region should be constructed. Typically, the full  feature representation of a region is formed by connecting the oriented gradient histograms from its overlapped sub areas, as showed in the Fig.\ref{fig:HOG_Gen}
\subsection{Build Integral Image for HOG}
To apply the integral image method to the calculation of the HOG features, first we recall how the histogram for each area inside a region is computed. Let the number of bins be $N_B$ (e.g. 9 for the Fig.\ref{fig:HOG_Gen}). Inside an area, the number of bins with same values are counted and grouped to form a histogram, the resulting histogram has the length equal to the number of bins. Some bins are empty (have zero value), such as the $0th,2nd,3rd$ and $8th$ bins for the first histogram in the Fig.\ref{fig:HOG_Gen}.

\begin{figure}[htb]
	\begin{center}
		\includegraphics[scale=0.5]{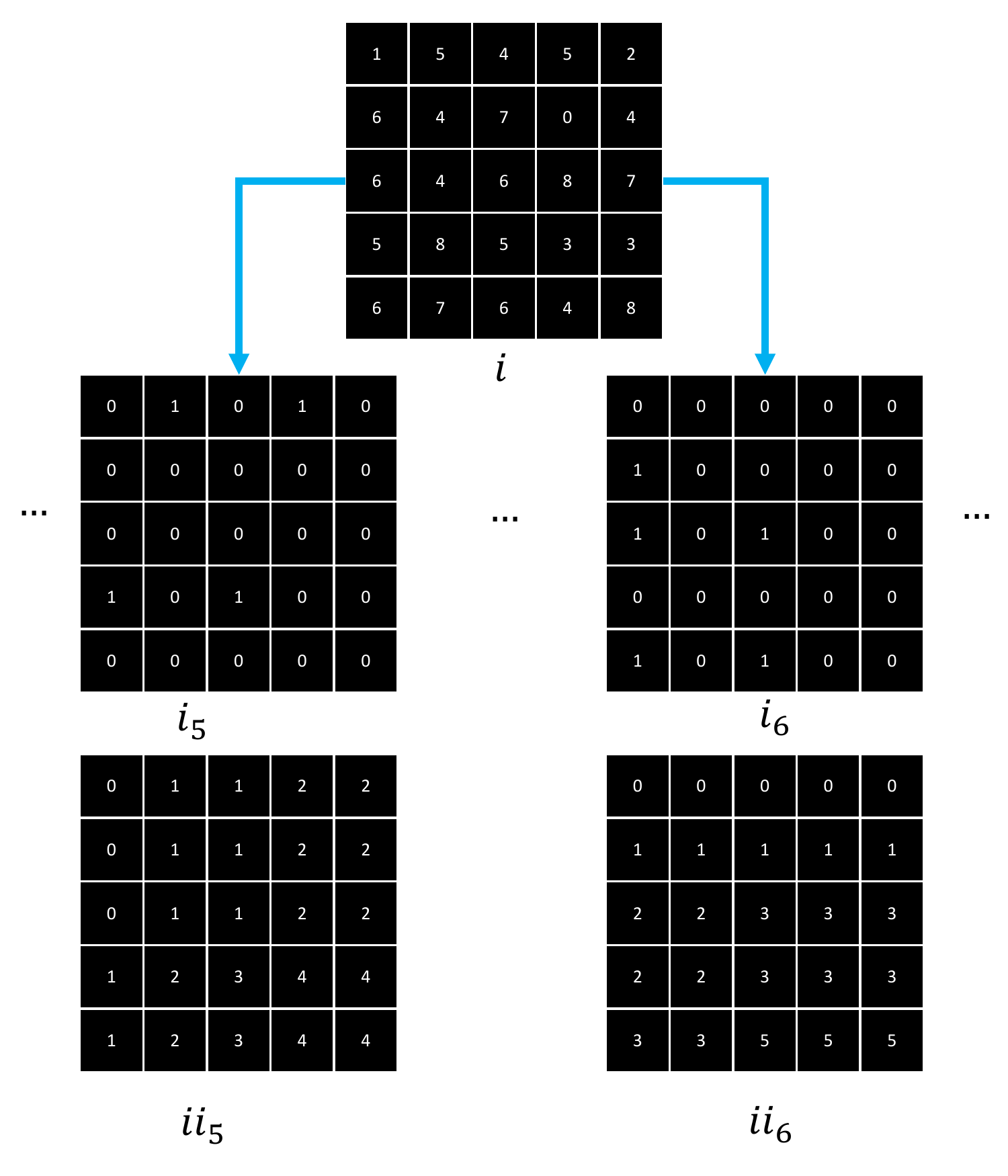}
	\end{center}
	\caption[Generation of Integral Images for HOG]{The steps of generating integral images for the oriented gradients map. The first row is the oriented gradients map generated in Sec.2. Decomposition of the map into $N_B$ images(here only shows the $5th$ and $6th$ ones). The integral image for each image in the second row is generated in the third row.}
	\label{fig:iiHOG}
\end{figure}

By observation, in order to speed up counting the bin values, we can have $N_B$ integral images, with different bin value corresponds to one of them. For the $n_{th}$ integral image, its value at point $(x,y)$ is equal to one when the value of the image  $i$ at that point has value equal to $n$. i.e.
\begin{equation}
i_n(x,y) = \left\{
\begin{array}{ll}
1, 			\ \ \ \ if\ \ i(x,y) = n\\
0, 			\ \ \ \ otherwise
\end{array}
\right.
\label{eq:ii_HOG}
\end{equation}
\indent The Fig.\ref{fig:iiHOG} shows how these integral images of an oriented gradients map are built. The image in the first row is decomposed into $N_B$ images(the second row) with value at different points given by Eq.\ref{eq:ii_HOG}. Then the integral over each image is carried out(Eq.\ref{eq:ii_cal_n}) and we get $N_B$ integral images for every bin value(the third row).
\begin{equation}
ii_n(x,y)=\sum_{x'\leq x,y'\leq y}i_n(x',y')
\label{eq:ii_cal_n}
\end{equation}

\subsection{Generate Histogram from Integral Images}
To compute the full histogram for a region(e.g. Fig.\ref{fig:HOG_Gen}), integral images generated in last sub section can be used to get the histogram value for each bin $B$ in different area $R$, denoted as $V_R^B$:
\begin{equation}
\begin{split}
V_R^B = \{ii_B(4) + ii_B(1) - ii_B(2)-ii_B(3)\\
\ points\ 1 \sim 4\ are\ indicated\ in\ the\ Fig. \ref{fig:integralimage} \}
\end{split}
\label{eq:ii_Hist}
\end{equation}

Using the Eq. \ref{eq:ii_Hist} to compute a full histogram with length $N$ only takes $N$ addition and $2N$ subtraction operations, i.e. with time complex $O(N)$. Once the integral image for each bin value is built, the time taken to construct the full HOG features is independent of the size of the input image. In a practical implementation, many areas for a given image will be visited more than once, the histograms for all those area can be cached first before evaluation is carried out. \\
The afterward steps, such as dataset preparation, training a classifier and  object detection using a HOG descriptor are not going to be repeated here, they can be found in other literatures or by reading some online tutorials.
\section{Conclusion}
This paper describes technologies can be used to accelerate the overall speed of object detection using the HOG descriptor. Two main modifications are made, one is the introduction of lookup table for computing the oriented gradients of an image. Another is the application of integral images for HOG features calculation. \\
\indent Other tips such as discard of the magnitude as mentioned in Sec.2, which will speed up the oriented gradients calculation. The selection of salient features using AdaBoost\cite{ratsch2001soft} or other methods\cite{xuefeng2014face}\cite{jolliffe2002principal} can reduce the dimensionality of the resulted features, which can speed up the training procedure and the inner product calculation during objectness evaluation. For some embed systems, such as application of a digital signal processor, if the input image is small enough(e.g. $352\times 288$ YUV videos from some cameras) that the usage 16-bit data types for integral images will not overflow, the speed can be doubled if codes are properly engineered.\\
\indent In a common personal desktop computer with a Intel i5 CPU, the modified HOG descriptor (my implementation) takes about 30 ms to detect objects for an image of size $800\times600$. If the code is built to 64-bit version, it only takes about 15 ms if the source code is properly engineered. If the size of the input image is $352\times 288$, it takes about 8 ms, or less than 5 ms for 64-bit version.\\
\section{Acknowledgments}
This paper is a brief review of the research I had done many years ago, specially thanks to Jixiang Huang for all the very helpful discussion during the time we worked together. \\
\indent For more information, please check the online video on  the application of the modified HOG descriptor for vehicle detection \url{https://www.youtube.com/watch?v=ge3HPE_qHCc}

{\small
\bibliographystyle{ieee}
\bibliography{egbib}

\begin{thebibliography}{1}\itemsep=-1pt

\bibitem{bay2006surf}
H.~Bay, T.~Tuytelaars, and L.~Van~Gool.
\newblock Surf: Speeded up robust features.
\newblock In {\em European conference on computer vision}, pages 404--417.
  Springer, 2006.

\bibitem{crow1984summed}
F.~C. Crow.
\newblock Summed-area tables for texture mapping.
\newblock {\em ACM SIGGRAPH computer graphics}, 18(3):207--212, 1984.

\bibitem{dalal2005histograms}
N.~Dalal and B.~Triggs.
\newblock Histograms of oriented gradients for human detection.
\newblock In {\em Computer Vision and Pattern Recognition, 2005. CVPR 2005.
  IEEE Computer Society Conference on}, volume~1, pages 886--893. IEEE, 2005.

\bibitem{huang2009fast}
C.~Huang, K.~Lin, and F.~Long.
\newblock A fast eye localization algorithm using integral image.
\newblock In {\em Computational Intelligence and Design, 2009. ISCID'09. Second
  International Symposium on}, volume~1, pages 231--234. IEEE, 2009.

\bibitem{jolliffe2002principal}
I.~Jolliffe.
\newblock {\em Principal component analysis}.
\newblock Wiley Online Library, 2002.

\bibitem{ratsch2001soft}
G.~R{\"a}tsch, T.~Onoda, and K.-R. M{\"u}ller.
\newblock Soft margins for adaboost.
\newblock {\em Machine learning}, 42(3):287--320, 2001.

\bibitem{viola2001rapid}
P.~Viola and M.~Jones.
\newblock Rapid object detection using a boosted cascade of simple features.
\newblock In {\em Computer Vision and Pattern Recognition, 2001. CVPR 2001.
  Proceedings of the 2001 IEEE Computer Society Conference on}, volume~1, pages
  I--I. IEEE, 2001.

\bibitem{viola2004robust}
P.~Viola and M.~J. Jones.
\newblock Robust real-time face detection.
\newblock {\em International journal of computer vision}, 57(2):137--154, 2004.

\bibitem{xuefeng2014face}
C.~Xuefeng, L.~Fei, and C.~Huang.
\newblock Face recognition by zero-ratio based lgbp features.
\newblock In {\em Intelligent Control and Automation (WCICA), 2014 11th World
  Congress on}, pages 5605--5608. IEEE, 2014.

\end{thebibliography}


@inproceedings{dalal2005histograms,
	title={Histograms of oriented gradients for human detection},
	author={Dalal, Navneet and Triggs, Bill},
	booktitle={Computer Vision and Pattern Recognition, 2005. CVPR 2005. IEEE Computer Society Conference on},
	volume={1},
	pages={886--893},
	year={2005},
	organization={IEEE}
}
@article{crow1984summed,
	title={Summed-area tables for texture mapping},
	author={Crow, Franklin C},
	journal={ACM SIGGRAPH computer graphics},
	volume={18},
	number={3},
	pages={207--212},
	year={1984},
	publisher={ACM}
}

@inproceedings{huang2009fast,
	title={A fast eye localization algorithm using integral image},
	author={Huang, Chunde and Lin, Kunhui and Long, Fei},
	booktitle={Computational Intelligence and Design, 2009. ISCID'09. Second International Symposium on},
	volume={1},
	pages={231--234},
	year={2009},
	organization={IEEE}
}
@inproceedings{viola2001rapid,
	title={Rapid object detection using a boosted cascade of simple features},
	author={Viola, Paul and Jones, Michael},
	booktitle={Computer Vision and Pattern Recognition, 2001. CVPR 2001. Proceedings of the 2001 IEEE Computer Society Conference on},
	volume={1},
	pages={I--I},
	year={2001},
	organization={IEEE}
}
@inproceedings{bay2006surf,
	title={Surf: Speeded up robust features},
	author={Bay, Herbert and Tuytelaars, Tinne and Van Gool, Luc},
	booktitle={European conference on computer vision},
	pages={404--417},
	year={2006},
	organization={Springer}
}
@article{viola2004robust,
	title={Robust real-time face detection},
	author={Viola, Paul and Jones, Michael J},
	journal={International journal of computer vision},
	volume={57},
	number={2},
	pages={137--154},
	year={2004},
	publisher={Springer}
}
@article{ratsch2001soft,
	title={Soft margins for AdaBoost},
	author={R{\"a}tsch, Gunnar and Onoda, Takashi and M{\"u}ller, K-R},
	journal={Machine learning},
	volume={42},
	number={3},
	pages={287--320},
	year={2001},
	publisher={Springer}
}
@inproceedings{xuefeng2014face,
	title={Face recognition by Zero-Ratio based LGBP features},
	author={Xuefeng, Cheng and Fei, Long and Huang, Chunde},
	booktitle={Intelligent Control and Automation (WCICA), 2014 11th World Congress on},
	pages={5605--5608},
	year={2014},
	organization={IEEE}
}
@book{jolliffe2002principal,
	title={Principal component analysis},
	author={Jolliffe, Ian},
	year={2002},
	publisher={Wiley Online Library}
}

}

\end{document}